%% file: main.tex
\documentclass[10pt,twocolumn,letterpaper]{article}

\usepackage{cvpr}              

\input{preamble}

\definecolor{cvprblue}{rgb}{0.21,0.49,0.74}
\usepackage[pagebackref,breaklinks,colorlinks,citecolor=cvprblue]{hyperref}

\title{Leveraging Positional Encoding for Robust Multi-Reference-Based \\ Object 6D Pose Estimation}

\author{Jaewoo Park \and Jaeguk Kim \and Nam Ik Cho \and \
Department of ECE, INMC, Seoul National University, Seoul, Korea\\
{\tt\small \{bjw0611, jaeguk, nicho\}@snu.ac.kr}
}

\begin{document}
\maketitle
\input{sec/0_abstract}    
\input{sec/1_intro}
\input{sec/2_related}
\input{sec/3_method}

\input{sec/4_experiment}
\input{sec/5_conclusion}
{
    \small
    \bibliographystyle{ieeenat_fullname}
    \bibliography{main}
}


\end{document}

%% file: preamble.tex
\usepackage{graphicx}
\usepackage{amsmath}
\usepackage{amssymb}
\usepackage{booktabs}
\usepackage{multirow}
\usepackage[ruled,vlined]{algorithm2e}
\DontPrintSemicolon
\SetAlCapSkip{.5\baselineskip}
\SetCommentSty{commentFont}

\usepackage{subcaption}
\usepackage[dvipsnames]{xcolor}


%% file: sec/0_abstract.tex
\begin{abstract}
Accurately estimating the pose of an object is a crucial task in computer vision and robotics. There are two main deep learning approaches for this: geometric representation regression and iterative refinement. However, these methods have some limitations that reduce their effectiveness. In this paper, we analyze these limitations and propose new strategies to overcome them. To tackle the issue of blurry geometric representation, we use positional encoding with high-frequency components for the object's 3D coordinates. To address the local minimum problem in refinement methods, we introduce a normalized image plane-based multi-reference refinement strategy that's independent of intrinsic matrix constraints. Lastly, we utilize adaptive instance normalization and a simple occlusion augmentation method to help our model concentrate on the target object. Our experiments on Linemod, Linemod-Occlusion, and YCB-Video datasets demonstrate that our approach outperforms existing methods. We will soon release the code.
\end{abstract}

%% file: sec/1_intro.tex
\section{Introduction}
\label{sec:intro}
Estimating the 6D pose of an object in a monocular image is crucial for interactions in both physical and virtual worlds, and has a wide range of uses such as augmented reality \cite{marchand2015pose}, autonomous driving \cite{manhardt2019roi,chen2017multi}, robotic grasping \cite{collet2011moped,wang2019densefusion}, and more. Several research studies have been conducted over the years to enhance object pose estimation and meet the demands of these applications \cite{sundermeyer2023bop}.

\begin{figure}[t]
  \centering
  \includegraphics[width=1.0\linewidth]{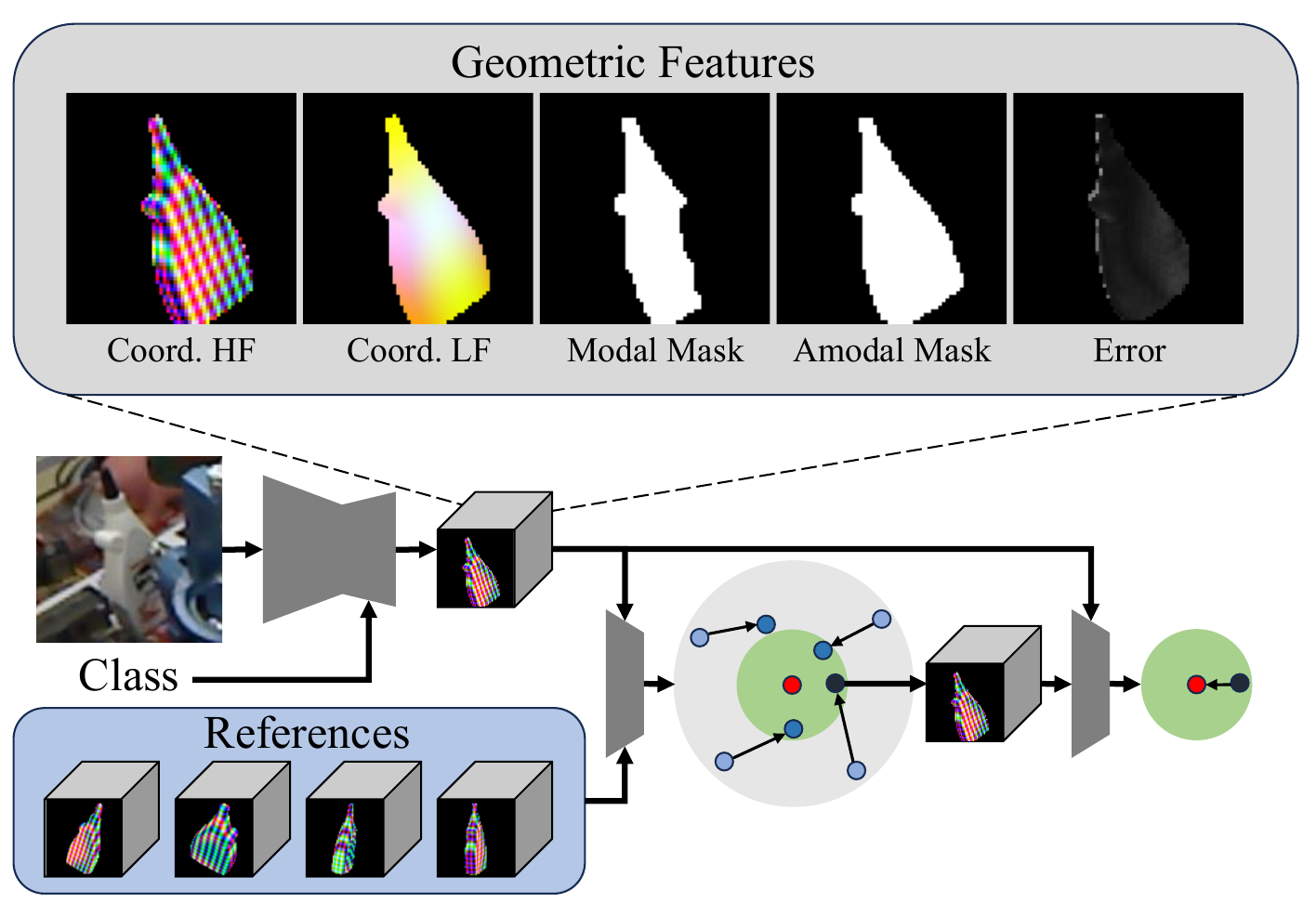}
   \caption{\textbf{Key idea of our method:} Our model estimates positional encoding, mask, and confidence of the query object and utilizes multiple references for reliable refinement estimations.}
   \label{fig:key_idea}
\end{figure}

\begin{figure}
  \centering
  \begin{subfigure}{0.68\linewidth}
    \includegraphics[width=1.0\linewidth]{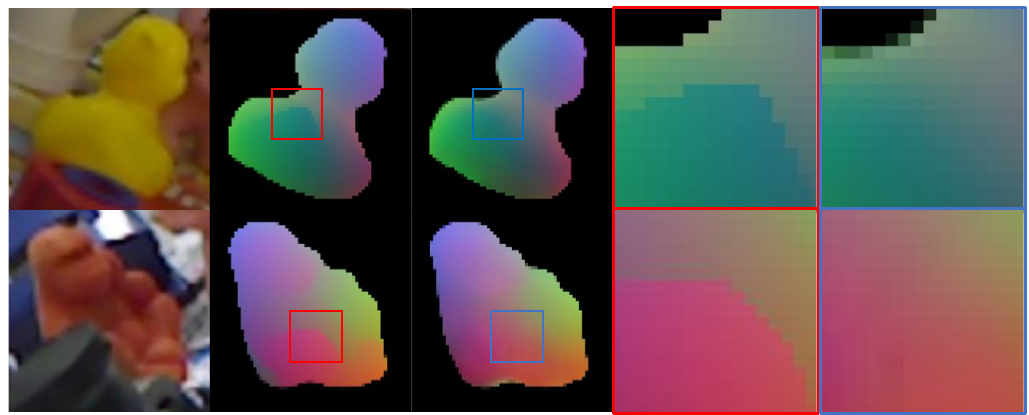}
    \subcaption{}
  \end{subfigure}
  \hfill
  \begin{subfigure}{0.28\linewidth}
    \includegraphics[width=1.0\linewidth]{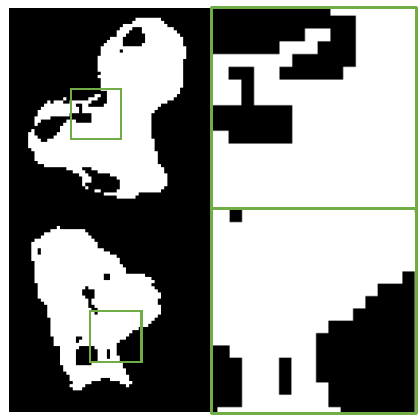}
    \subcaption{}
  \end{subfigure}
  \caption{\textbf{Failure cases in coordinate estimation:} (a) We visualize the blurry 2D-3D coordinate estimation examples. The red box represents ground-truths and the blue box represents predictions. (b) We also visualize inlier samples selected from the RANSAC method, which fails to cut out blurry estimations.}
  \label{fig:motivation_coord}
\end{figure}

There are two main approaches to the object pose estimation. The first approach estimates the 2D-3D coordinates of the object surface and then uses PnP-RANSAC \cite{park2019pix2pose, hu2022perspective} or an additional neural network head \cite{li2019cdpn, wang2021gdr, di2021so} for pose estimation. However, these methods face difficulties when dealing with round and textureless objects, which are made up of smooth geometric representations. As a result, these objects often cause blurry estimations, making the model hard to distinguish discontinuous coordinates and cause PnP-RANSAC to estimate the pose imprecisely as shown in Fig. \ref{fig:motivation_coord}. Even methods that use direct pose regression heads, such as those found in \cite{wang2021gdr} and \cite{di2021so}, still struggle with this problem. To overcome this limitation, approaches like \cite{su2022zebrapose} use discrete external texture. However, they only represent the object with the over-segmented pieces rather than representing the overall object shape in their representation.

\begin{figure}[t]
  \centering
  \includegraphics[width=1.0\linewidth]{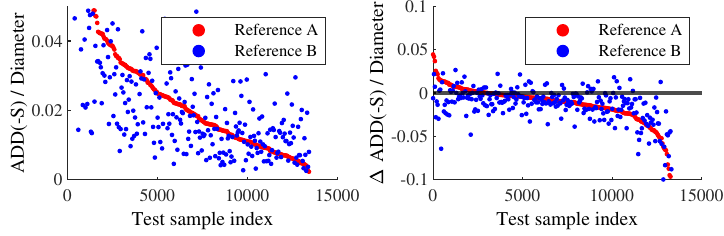}
   \caption{\textbf{Different patterns depending on the reference:} We visualize final estimations and improvements of render-and-compare models for each test sample based on two different references.}
   \label{fig:motivation_local}
\end{figure}

On the other hand, several recent methods \cite{li2018deepim, labbe2020cosypose, park2022dprost, hai2023shape} have employed the iterative render-and-compare approach to estimate the relative pose between a reference and a query image. These approaches progressively target finer search spaces for more precise estimation. However, this limited search space makes them difficult to refine a pose that gets stuck in a local minimum during the initial iteration. One possible solution is to use a bagging ensemble \cite{ganaie2022ensemble}, which employs multiple initial references, enabling the system to find reliable refinement among them. As shown in Fig. \ref{fig:motivation_local}, the estimation of each test sample varies with each reference-based refinement. Hence, this approach effectively mitigates the issue of local minima. However, refinement-based methods are based on a relative pose parametrization, which is only feasible when the zoomed-in intrinsic matrix for both the query and the reference are identical \cite{li2018deepim}. Therefore, online rendering of the reference is necessary to align with each query's intrinsic matrix. This can become computationally demanding when extended to multiple references.

Considering the advantages and limitations of previous methods, we introduce a \textbf{M}ulti-Reference \textbf{R}efinement method with \textbf{P}ositional \textbf{E}ncoding (MRPE) for pose estimation as shown in Fig. \ref{fig:key_idea}. Inspired from structure light scanning \cite{rocchini2001low} and Neural Radiance Fields (NeRF) \cite{mildenhall2020nerf}, we design our method to estimate the positional encoding of the normalized object coordinates, along with the mask and prediction error. The positional encoding method provides comprehensive shape information, similar to contour lines, which enables the model to focus on shape details via high-frequency components. This significantly reduces blurry estimations in challenging objects.

In the next step, we apply the render-and-compare framework in the geometric domain. Specifically, the estimated positional encoding is compared with the offline multi-references for relative pose estimation. To achieve this, we introduce an intrinsic matrix untangled pose update method. This improves performance by using a bagging ensemble without slowing down due to the bottleneck of rendering multiple references.

Furthermore, to address objects with severe occlusion, we employ the Adaptive Instance Normalization (AdaIN) \cite{karras2019style,huang2017arbitrary} technique to adapt the model to focus on the target object according to the given object class. Experimental results on the Linemod (LM), Linemod-Occlusion (LM-O), and YCB-Video (YCB-V) datasets indicate that our approach generally outperforms existing state-of-the-art methods.

Our contributions are as follows:
\begin{itemize}
\item We address the blurry estimation issue by utilizing positional encoding for object coordinates as the target of the estimation.
\item We introduce a pose update module that doesn't rely on intrinsic matrices. This module enables offline multi-references in the render-and-compare process, which eliminates rendering bottlenecks and reduces local minimum issues.
\item We present an AdaIN-based model that can focus on target objects even when they are severely occluded.
\end{itemize}

%% file: sec/2_related.tex
\section{Related Work}
\label{sec:related}
Deep learning-based models have shown impressive capabilities in extracting high-level features from images \cite{he2016deep, ren2015faster}. As a result, several pose estimation methods have utilized these models to estimate the geometric features of the target object. Typically, these approaches identify sparse keypoints \cite{wu2022vote, peng2019pvnet, song2020hybridpose}, bounding box corners \cite{hu2019segmentation, rad2017bb8}, dense 2D-3D correspondence \cite{park2019pix2pose, hu2022perspective}, or UV maps \cite{zakharov2019dpod}, which are then used in PnP-RANSAC \cite{lepetit2009epnp} for pose estimation. Although these intermediate representations may help the model understand the shape of the objects, direct pose estimation from these representations is vulnerable to misalignment of the predictions.

\begin{figure*}[t]
  \centering
   \includegraphics[width=1.0\linewidth]{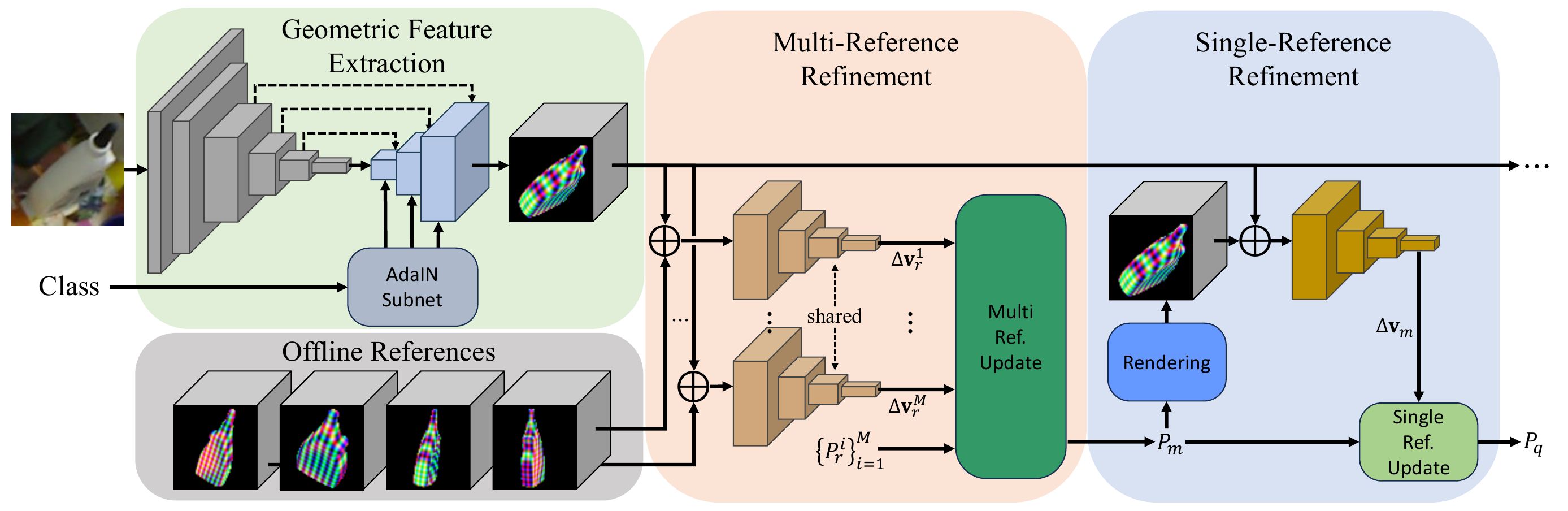}
   \caption{\textbf{Overview of our method:} Our approach follows a four-step process. Firstly, we generate offline references. Then, for a query image, we estimate its geometric features and determine its relative pose with respect to the references. After that, we identify a reliable pose and refine it using the standard render-and-compare strategy.}
   \label{fig:overview}
\end{figure*}

On the other hand, some methods have been proposed to refine an object's pose iteratively. For instance, \cite{li2018deepim} and \cite{labbe2020cosypose} use a rendered mesh as a reference, while \cite{park2022dprost} incorporates projected space-carved features, and \cite{hai2023shape} applies rigidity constraints to iterative optical flow. However, all these approaches rely on RGB domain features for both queries and references, which can limit their performance, especially when dealing with textureless objects or significant occlusions where RGB features may not provide sufficient information. To overcome these limitations, our method begins by estimating the rich geometric domain features of the query image. We then combine these features with a rigid reference in the geometric domain to accurately estimate the relative pose.

As the NeRF \cite{mildenhall2020nerf} is actively being researched recently, there have been methods that apply NeRF to 6D pose estimation. For instance, iNeRF \cite{yen2021inerf} utilized backpropagation through a trained NeRF model to estimate the pose of non-occluded, background-free objects, while TexPose \cite{chen2023texpose} introduced a method for environments where object labels are limited. However, NeRF models are known take a lot of time to train, and their pixel-wise inference can also cause bottlenecks. To avoid these issues, instead of training a NeRF model, we decided to use only positional encoding, which has been suggested as a way to enhance the performance of NeRF models \cite{mildenhall2020nerf, tancik2020fourier}. Note that Uni6D \cite{jiang2022uni6d} used positional encoding on the 2D image space coordinates $(u,v)$ as an additional input, following the approach proposed in Transformer \cite{vaswani2017attention, dosovitskiy2020image} papers. However, our approach is distinct from theirs, as we focus on estimating the 3D object space coordinates by applying NeRF-style positional encoding.

%% file: sec/3_method.tex
\section{Method}
\label{sec:method}
This section presents our method to estimate the pose $P_{q}$ of a 3D model $\mathcal{M}$ in a given query image. Similar to \cite{li2019cdpn,labbe2020cosypose,wang2021gdr,park2022dprost}, we utilize an off-the-shelf detector to zoom-in on the object image $I_{q}$ and obtain intrinsic matrix $K_{q}$. 

As illustrated in Fig. \ref{fig:overview}, our model is divided into four distinct modules.

\noindent \textbf{Offline Reference Generation:} Before beginning the training process, we choose $M$ representative samples for each object from the training set to use as references. For each reference, we generate geometric features $\{F_{r}^{i}\}_{i=1}^{M}$, which are used during both training and inference phases. Please refer to Section \ref{Sec:offline_reference_generation} for a detailed explanation of this process.

\noindent \textbf{Geometric Feature Extraction:} In this module, a detailed query feature $F_{q}$ is extracted via the feature extractor. To help the model focus on the target object, we use adaptive instance normalization (AdaIN) \cite{karras2019style,huang2017arbitrary} with detected object class. We provide further details of the model in Section \ref{Sec:geometric_feature_extraction}. 

\noindent \textbf{Multi-Reference Refinement:} The pose prediction head uses $F_{q}$ and each of $F_{r}^{i}$ to calculate the refinement parameter $\Delta \mathbf{v}_{r}^{i}$, which includes the rotational and positional differences between the query and the $i$-th reference. The $i$-th reference pose $P_{r}^{i}$ is then updated with $\Delta \mathbf{v}_{r}^{i}$ to generate a pose candidate $P_{c}^{i}$ for the query. Subsequently, the most reliable candidate is selected as the initial pose estimation $P_{m}$.

\noindent \textbf{Single-Reference Refinement:} Finally, we generate the feature $F_{m}$ of $P_{m}$ and compare it with $F_{q}$ to determine the refinement parameter $\Delta \mathbf{v}_{m}$. Using this parameter, we update $P_{m}$ to obtain the final estimation, $P_{q}$ iteratively. For more details on both stages of refinement, please refer to Section \ref{Sec:pose_refinement}.

Also, as the training dataset has a limited number of occluded samples, we introduce a simple occlusion augmentation approach. This approach superimposes the mask and image of another object from the same mini-batch, thus creating additional occlusions during training.

\subsection{Offline Reference Generation}
\label{Sec:offline_reference_generation}
To choose a diverse set of reference views of an object from the training set we use the farthest point sampling (FPS) algorithm as described in \cite{park2022dprost}. The selected references' pose $\{P_{r}^{i}\}_{i=1}^{M}$ is used to project and generate a 2D-3D correspondence map in geometric features with the positional encoding function. This is done as follows: 
\begin{gather}
    G_{r}^{i} = \gamma(\pi(K_{r}^{i}, P_{r}^{i}, \mathcal{M}))
    \label{equ:geometric}
\end{gather}
Here, $G_{r}^{i} \in \mathbb{R}^{h \times w \times 6N}$ is an encoded feature, $\gamma(\cdot)$ denotes the positional encoding function with $N$ frequencies as presented in \cite{mildenhall2020nerf}, and $\pi(\cdot)$ represents projection function with the intrinsic matrix $K_{r}^{i}$, pose $P_{r}^{i}$, and the 3D model coordinate $\mathcal{M}$. 

$G_{r}^{i}$ is then concatenated with reference image $I_{r}^{i}$, modal mask $M_{r^{-}}^{i}$, and amodal mask $M_{r^{+}}^{i}$ to generate the overall reference feature $F_{r}^{i} \in \mathbb{R}^{h \times w \times (6N+5)}$, and then used in multi-reference refinement module in both the training and test phases. 

We have also conducted experiments for the mesh-less scenario, where we used the surface coordinate of the space-carved voxel in place of $\mathcal{M}$ as suggested in \cite{park2022dprost}.

\subsection{Geometric Feature Extraction}
\label{Sec:geometric_feature_extraction}
Our feature extractor is based on a convolutional neural network (CNN) encoder-decoder architecture which is designed to estimate the dense feature of $I_{q}$. The model estimates the geometric feature $G_{q} \in \mathbb{R}^{h \times w \times 6N}$, similar to Equation \ref{equ:geometric}. In addition to this, the model also predicts the error $E_{q} \in \mathbb{R}^{h \times w \times 6N}$ of the geometric feature estimation itself. The model also predicts amodal mask $M_{q^{+}}$ and modal mask $M_{q^{-}}$, which are then concatenated with $I_{q}$ to generate query feature $F_{q} \in \mathbb{R}^{h \times w \times 12N+5}$.

The $E_{q}$ is designed to handle the inherent geometric ambiguity of symmetric objects. Since the positional encoding $\gamma(\cdot)$ incorporates both symmetrical sine encoding and asymmetrical cosine encoding, $E_{q}$ is used as an indicator to inform the model to only focus on cosine embeddings for symmetric objects, and both sine and cosine embeddings for asymmetric objects.

In addition, another source of ambiguity arises when training a single model for multiple target objects. Previous methods did not utilize the detected object class in the feature extractor, making it difficult for the model to distinguish the target object when multiple objects were overlapped. To address this issue, we implemented the AdaIN \cite{huang2017arbitrary} method to help the feature extractor focus on the target object. Specifically, the detected class of the object is embedded through a subnetwork based on Fully Connected (FC) layers, and then incorporated into the feature extractor. The following equation represents how the AdaIN method works:
\begin{gather}
AdaIN(X_{i}, \alpha_{i}, \beta_{i}) = \alpha_{i} \frac{X_{i}-\mu(X_{i})}{\sigma(X_{i})} +\beta_{i},
\end{gather}
Here, $X_{i}$ denotes the $i$-th layer feature in the feature extractor head, $\alpha_{i}$ and $\beta_{i}$ are the parameters estimated by the AdaIN subnetwork, and $\mu(\cdot)$ and $\sigma(\cdot)$ denote the instance-wise mean and standard deviation functions, respectively.

\subsection{Pose Refinement}
\label{Sec:pose_refinement}

\begin{algorithm}[t]
\caption{Untangled pose refinement}
\SetKwInOut{Input}{input}
\SetKwInOut{Output}{output}
\SetAlgoLined
\Input{Predictions: $\mathcal{V} = \{\mathbf{v}_{\text{rot}}^{i}, \mathbf{v}_{xyz}^{i}\}_{i=1}^{N}$, \\
      Ref. parameters: $\mathcal{R} = \{f_{r}^{i}, K_{r}^{i}, R_{r}^{i}, \mathbf{t}_{r}^{i}\}_{i=1}^{N}$, \\
      Que. parameter: $f_{q}, K_{q}$}
\Output{Refined pose: $\{R_{c}^{i}, \mathbf{t}_{c}^{i}\}_{i=1}^{N}$}
\ForEach {$i = 1$ \KwTo $N$}{
\tcp{refined translation z} 
$t_{cz}^{i} \gets t_{rz}^{i}(v_{z}^{i} + 1) f_{q} / f_{r}^{i}$ \\
\tcp{normalized plane reference center}
$\mathbf{o}_{r} \gets K_{r}^{i} \mathbf{t}_{r}^{i} / t_{rz}^{i}$ \\
\tcp{refined object center}
$\mathbf{o}_{c} \gets \mathbf{o}_{r} + \left( v_{x}^{i}, v_{y}^{i}, 0 \right)^T$ \\
\tcp{normalized plane to query camera space}
$\mathbf{t}_{cxy}^{i} \gets t_{cz}^{i}\left( {(K_{q}})^{-1} \mathbf{o}_{c} \right)_{xy}$ \\
$R_{r, a}^{i} \gets$ Convert to allocentric from $R_{r}^{i}, \mathbf{t}_{r}^{i}$\\
$R_{\text{rot}}^{i} \gets$ Convert to matrix from 6d vector $\mathbf{v}_{\text{rot}}^{i}$ \\
$R_{c, a}^{i} \gets R_{rot}^{i} R_{r, a}^{i}$ \tcp*{allocentric refinement}
$R_{c}^{i} \gets$ Convert to egocentric from $R_{c, a}^{i}, \mathbf{t}_{c}^{i}$\\
}
\label{alg:FDP_refinement}
\end{algorithm}

\noindent \textbf{Multi-Reference Refinement:}
To refine the parameters between the query and references, we concatenate the feature representations $F_{q}$ and each $F_{r}^{i}$ and feed them to a CNN-based pose estimation head. The head's final layer has a branch for a six-dimensional rotation representation $\mathbf{v}_{\text{rot}}^{i} \in \mathbb{R}^{6}$ output, as well as a branch for normalized image plane translation and scale $\mathbf{v}_{xyz}^{i} \in \mathbb{R}^{3}$.

Algorithm \ref{alg:FDP_refinement} updates the reference pose to a query pose candidate $P_{c}^{i}$. To account for the scale difference due to focal length, we first update the $z$ value of the translation vector $t_{cz}^{i}$. Next, we project the object center of reference onto the normalized image plane to separate the scale and offset issues of the intrinsic matrix $K_{r}^{i}$. We then update the projected object center $\mathbf{o}_{r}$ of each reference by estimated $v_{x}^{i}$ and $v_{y}^{i}$, and the refined centers $\mathbf{o}_{c}$ are unprojected using the given query intrinsic matrix $K_{q}$ and scaled by $t_{cz}^{i}$ to find $x$ and $y$ values of the translation vector $\textbf{t}_{cxy}^{i}$. 

For rotation refinement, we follow the allocentric rotation refinement scheme proposed in \cite{kundu20183d}, which addresses viewpoint-invariant rotation. First, the rotation matrix $R_{r}^{i}$ is transformed into an allocentric representation $R_{r,a}$. Then, the relative rotation matrix $R_{\text{rot}}^{i}$ from predicted $\mathbf{v}_{\text{rot}}^{i}$ refines the rotation in the allocentric domain to $R_{c,a}^{i}$. Finally, the refined $R_{c, a}^{i}$ is transformed into the egocentric domain using the predicted $\mathbf{t}_{c}^{i}$ to obtain the rotation prediction. 

As a result, the estimated $\Delta \mathbf{v}^{i}_{r}$ are only utilized in the allocentric domain, which removes the confusion caused by the translation. Additionally, since the translation update is conducted on the normalized image plane, the estimated refinement parameters from the model are untangled from the intrinsic matrix. This helps the model learn the rule of the parameter efficiently. Moreover, different intrinsic matrix-based references can be used with the query image, allowing for offline-generated references to be used and significantly reducing the computational cost of generating the references for each query.

\noindent \textbf{Reliable Pose Voting:}
Inspired by the effect of the ensemble bagging, renowned for enhancing prediction reliability, we use the medoid values from $\{{R_{c}^{i}, \mathbf{t}_{c}^{i}}\}_{i=1}^{N}$, similar to a voting scheme. Specifically, the sum of distances between each $\mathbf{t}_{c}^{i}$ and the other candidates are used as a metric for translation. Similarly, the sum of geodesic distances between each $R_{c}^{i}$ and the other candidates are used as a metric for rotation. The process can be expressed as follows:
\begin{gather}
d_{\mathbf{t}}^{i} =  \sum_{k \neq i} \left\|\mathbf{t}_{c}^{k} - \mathbf{t}_{c}^{i}\right\|_{2}, \\
d_{R}^{i} = \sum_{k \neq i} \arccos{\left(\left(tr\left({R_{c}^{i}}^{T} R_{c}^{k}\right)-1\right)/2 \right)} 
\end{gather}
where $tr$ represents the trace function of a matrix. Subsequently, we find the medoid rotation $R_{m}$ and translation $\mathbf{t}_{m}$ by selecting each component with the minimum $d_{\mathbf{t}}^{i}$ and $d_{R}^{i}$, respectively.

\noindent \textbf{Single-Reference Refinement:}
We use the selected $P_{m}=[R_{m}|\mathbf{t}_{m}]$ and given $K_{q}$ to create a medoid reference's feature $F_{m} \in \mathbb{R}^{h \times w \times (6N+2)}$. This feature is similar to $F_{r}^{i}$, but without the image. We aggregate $F_{m}$ with $F_{q}$ and use it to estimate the refinement parameters $\Delta\mathbf{v}_{m}$ with the pose estimation head. Unlike in the initial multi-reference refinement, we do not use multi-reference to remove the bottleneck of rendering multiple geometric features. Since the intrinsic matrices are the same between $F_{m}$ and $F_{q}$, we use the plain pose update method, which is the same as in \cite{labbe2020cosypose, park2022dprost}, to find the final pose estimation $P_{q}$. We repeat this step for fine refinement and find that two steps are enough for saturation, based on experimental results.

\subsection{Objective Function}
Our model has two categories of objective functions: one for extracting the feature $F_{q}$ from the query and the other for estimating the pose $P_{c}^{i}$ and $P_{q}$. In the feature extraction process, we employ a masked $L_{1}$ loss for $G_{q}$ and $E_{q}$, which can be formulated as follows:
\begin{gather}
\mathcal{L}_{G} = \| \bar{M}_{q^{-}} \odot ( G_{q} - \bar{G}_{q} ) \|_{1}, \\
\mathcal{L}_{E} = \| \bar{M}_{q^{-}} \odot ( E_{q} - \|{G}_{q} - \bar{G}_{q} \|_{1} ) \|_{1}
\end{gather}
where $\bar{M}_{q^{-}}$ is the ground-truth modal mask of a query, $\bar{G}_{q}$ is ground-truth query geometric feature, and $\odot$ is an element-wise multiplication. Note that as $E_{q}$ targets estimating error of the geometric feature, $\|{G}_{q} - \bar{G}_{q} \|_{1}$ value is the ground-truth of the $E_{q}$. For modal mask $M_{q^{-}}$ and amodal mask $M_{q^{+}}$ predictions, we use simple $L_{1}$ loss as follows:
\begin{gather}
\mathcal{L}_{M} = \| \bar{M}_{q^{-}} - M_{q^{-}}\|_{1} + \| \bar{M}_{q^{+}} - M_{q^{+}}\|_{1}
\end{gather}
where $\bar{M}_{q^{+}}$ is the ground-truth amodal mask of a query image.

Regarding the objective function for predicting pose, we adopt the grid-matching loss and grid-distance loss function proposed in \cite{park2022dprost}, which can be formulated as follows:
\begin{gather}
\mathcal{L}_{P_{c}^{i}} = \|\mathcal{\bar{G}}_{q} - \mathcal{G}_{c}^{i}\|_{2} + \left\|\|\bar{\mathbf{t}}_{q}\|_{2} - \|\mathbf{t}_{c}^{i}\|_{2}\right\|_{1}
\label{equ:grid_loss}
\end{gather}
where $\mathcal{\bar{G}}_{q}$ is a grid generated with the ground-truth query pose $\bar{P}_{q}$, following the grid generation method in \cite{park2022dprost}, and $\bar{\mathbf{t}}_{q}$ is the ground-truth translation vector, and the $\mathcal{G}_{c}^{i}$ is the grid from the estimated pose $P_{c}^{i}$. We also use the same Equation \ref{equ:grid_loss} as an objective function $\mathcal{L}_{P_{q}}$ for $\mathcal{G}_{q}$ and $\mathbf{t}_{q}$ from $P_{q}$. Unlike the approach employed in \cite{park2022dprost}, where the grid is used in dense projection, we only utilize 4 $\times$ 4 $\times$ 2 grid points, as they are sufficient for loss calculation. For further details about generating $\mathcal{G}$, refer to \cite{park2022dprost}.

The final objective function of the overall model can be formulated as
\begin{gather}
\mathcal{L}= \lambda \left(\mathcal{L}_{G} + \mathcal{L}_E + \mathcal{L}_{M} \right) + \sum_{i} \mathcal{L}_{P_{r}^{i}} + \mathcal{L}_{P_{q}}
\end{gather}
where $\lambda$ serves as a balance factor, which is set to 20 in our method.

%% file: sec/4_experiment.tex
\begin{table*}[t]
{\resizebox{1.0\linewidth}{!}
{\begin{tabular}{c|ccccccccc|cc}
\toprule
Method        & \begin{tabular}[c]{@{}c@{}}PoseCNN\\ \cite{Xiang2018posecnn}\end{tabular} & \begin{tabular}[c]{@{}c@{}}HybridPose\\ \cite{song2020hybridpose}\end{tabular} & \begin{tabular}[c]{@{}c@{}}GDR-Net\\ \cite{wang2021gdr}\end{tabular} & \begin{tabular}[c]{@{}c@{}}SO-Pose\\ \cite{di2021so}\end{tabular} & \begin{tabular}[c]{@{}c@{}}DProST\\ \cite{park2022dprost}\end{tabular} & \begin{tabular}[c]{@{}c@{}}DeepIM\\ \cite{li2018deepim}\end{tabular} & \begin{tabular}[c]{@{}c@{}}RePose\\ \cite{iwase2021repose}\end{tabular} & \begin{tabular}[c]{@{}c@{}}RNNPose\\ \cite{xu2022rnnpose}\end{tabular} & \begin{tabular}[c]{@{}c@{}}SCFlow\\ \cite{hai2023shape} \end{tabular} & \begin{tabular}[c]{@{}c@{}}Ours\\ (Res34)\end{tabular} & \begin{tabular}[c]{@{}c@{}}Ours\\ (Convnext-T)\end{tabular} \\ \hline
Init. Pose    & -                                                        & -                                                           & -                                                        & -                                                        & -                                                       & PoseCNN                                                 & PoseCNN                                                 & PoseCNN                                                  & PoseCNN                                                 & -           & -                \\ \hline
ADD(-S) 0.02d & -                                                        & -                                                           & 35.5                                                     & 45.9                                                     & 51.6                                                    & -                                                       & -                                                       & 50.4                                                     & -                                                       & 65.5        & \textbf{69.6}    \\
ADD(-S) 0.05d & -                                                        & -                                                           & 76.3                                                     & 83.1                                                     & 87.8                                                    & -                                                       & -                                                       & 85.6                                                     & 90.9                                                    & 93.2        & \textbf{95.0}    \\
ADD(-S) 0.1d  & 62.7                                                     & 94.5                                                        & 93.7                                                     & 96.0                                                     & 98.3                                                    & 88.6                                                    & 96.1                                                    & 97.4                                                     & \textbf{99.3}                                           & 99.2        & \textbf{99.3}    \\
\bottomrule
\end{tabular}}\caption{\textbf{Comparison on the LM Dataset.} We present a comparison based on the ADD(-S) metrics for the LM dataset. Init. Pose denotes the requirement of an initial pose estimator.}
\label{table:LM_sota}}
\end{table*}

\begin{table*}[t]
{\resizebox{1.0\linewidth}{!}
{\begin{tabular}{c|cccccccc|cc}
\hline
Method       & \begin{tabular}[c]{@{}c@{}}PoseCNN\\ \cite{Xiang2018posecnn}\end{tabular} & \begin{tabular}[c]{@{}c@{}}GDR-Net\\ \cite{wang2021gdr}\end{tabular} & \begin{tabular}[c]{@{}c@{}}SO-Pose\\ \cite{di2021so}\end{tabular} & \begin{tabular}[c]{@{}c@{}}DProST\\ \cite{park2022dprost}\end{tabular} & \begin{tabular}[c]{@{}c@{}}DeepIM\\ \cite{li2018deepim}\end{tabular} & \begin{tabular}[c]{@{}c@{}}RePose\\ \cite{iwase2021repose}\end{tabular} & \begin{tabular}[c]{@{}c@{}}RNNPose\\ \cite{xu2022rnnpose}\end{tabular} & \begin{tabular}[c]{@{}c@{}}SCFlow\\ \cite{hai2023shape}\end{tabular} & \begin{tabular}[c]{@{}c@{}}Ours\\ (Res34)\end{tabular} & \begin{tabular}[c]{@{}c@{}}Ours\\ (Convnext-T)\end{tabular} \\ \hline
Init. Pose   & -                                                   & -                                                   & -                                                   & -                                                  & PoseCNN                                            & PoseCNN                                            & PoseCNN                                             & PoseCNN                                            & -                                                      & -                                                           \\ \hline
ADD(-S) 0.1d & 21.3                                                & 49.1                                                & 56.8                                                & 43.8                                               & 53.6                                               & 60.3                                               & 66.4                                                & 73.2                                               & 78.9                                                   & \textbf{86.1}                                               \\
ADD(-S) AUC  & 61.3                                                & 80.2                                                & 83.9                                                & 69.2                                               & 81.9                                               & 80.8                                               & 83.1                                                & \textbf{-}                                         & 86.9                                                   & \textbf{89.1}                                               \\ \hline
\end{tabular}}\caption{\textbf{Comparison on the YCB-V Dataset.} We present a comparison based on the ADD(-S) metrics for the YCB-V dataset. Init. Pose denotes the requirement of an initial pose estimator.}
\label{table:YCB-V_sota}}
\end{table*}

\begin{table}[t]
{\resizebox{1.0\columnwidth}{!}
{\begin{tabular}{c|cccc|cc}
\hline
Method      & \begin{tabular}[c]{@{}c@{}}PoseCNN\\ \cite{Xiang2018posecnn}\end{tabular} & \begin{tabular}[c]{@{}c@{}}RePose\\ \cite{iwase2021repose}\end{tabular} & \begin{tabular}[c]{@{}c@{}}RNNPose\\ \cite{xu2022rnnpose}\end{tabular} & \begin{tabular}[c]{@{}c@{}}SCFlow\\ \cite{hai2023shape}\end{tabular} & \begin{tabular}[c]{@{}c@{}}Ours\\ (Res34)\end{tabular} & \begin{tabular}[c]{@{}c@{}}Ours\\ (Convnext-T)\end{tabular} \\ \hline
Init. Pose  & -                                                   & PoseCNN                                            & PoseCNN                                             & PoseCNN                                            & -                                                      & -                                                           \\ \hline
ape         & 9.6                                                 & 31.1                                               & 37.2                                                & -                                                  & \textbf{55.7}                                          & 48.1                                                        \\
can         & 45.2                                                & 80                                                 & 88.1                                                & -                                                  & \textbf{91.8}                                          & 89.5                                                        \\
cat         & 0.9                                                 & 25.6                                               & 29.2                                                & -                                                  & 49.1                                                   & \textbf{50.9}                                               \\
driller     & 41.4                                                & 73.1                                               & 88.1                                                & -                                                  & \textbf{89.6}                                          & 88.8                                                        \\
duck        & 19.6                                                & 43                                                 & 49.2                                                & -                                                  & 49.2                                                   & \textbf{52.9}                                               \\
eggbox*     & 22.0                                                & 51.7                                               & \textbf{67.0}                                       & -                                                  & 63.3                                                   & 57.5                                                        \\
glue*       & 38.5                                                & 54.3                                               & 63.8                                                & -                                                  & \textbf{77.4}                                          & 77.2                                                        \\
holepuncher & 22.1                                                & 53.6                                               & 62.8                                                & -                                                  & 78.7                                                   & \textbf{86.4}                                               \\ \hline
Average     & 24.9                                                & 51.6                                               & 60.7                                                & 67.0                                               & \textbf{69.4}                                          & 68.9                                                        \\ \hline
\end{tabular}}\caption{\textbf{Comparison on the LM-O Dataset.} We present a comparison based on the ADD(-S) 0.1d metric for each object in the LM-O dataset. Init. Pose denotes the requirement of an initial pose estimator.}
\label{table:LM-O_sota}}
\end{table}

\section{Experiment}
\label{sec:experiment}
\subsection{Experimental Setup}
\noindent \textbf{Datasets:} Our method is evaluated on Linemod (LM), Linemod-Occlusion (LM-O), and YCB-Video (YCB-V) datasets. The LM dataset includes 13 different objects, each with 1.2k annotated real images. Following the setting of \cite{brachmann2016uncertainty}, 15$\%$ of the total images are used for training the model, and the rest are used as the test set. The LM-O dataset is a subset of the LM dataset, containing 8 objects. Contrary to the LM dataset, where the target object is clearly visible at the image, the target objects in the LM-O test set are severely occluded by other objects, which makes it very challenging to address. Lastly, the YCB-V dataset comprises 21 objects in 92 video sequences recorded under various conditions. Consistent with previous methods, we use 80 sequences as the training set and the remaining 12 sequences as the test set. Additionally, we use the physically-based rendering (pbr) dataset provided by BOP challenge \cite{sundermeyer2023bop} along with real image datasets across all three datasets.

\noindent \textbf{Metrics:} We use the ADD(-S) score \cite{Xiang2018posecnn} as the main metric to evaluate the accuracy of our 6D pose estimation method. This is a common practice in previous research. ADD(-S) measures the average distance between the vertices of the object, which have been transformed with the estimated pose, and the ground-truth pose. If the object is symmetric, the measurement is taken between the nearest points instead. We basically use a threshold of 0.1 times the object's diameter to determine if a prediction is accurate. For the LM dataset, which is highly precise, we also report accuracy based on thresholds of 0.05 and 0.02 times the object's diameter. For the YCB-V dataset, we additionally report the Area Under Curve (AUC) of ADD(-S) up to a maximum of 10cm, consistent with previous research.

Furthermore, we compared our method against three challenge evaluation metrics in light of the recent BOP challenge: Maximum Symmetry-aware Projection Distance (MSPD), Maximum Symmetry-aware Surface Distance (MSSD), and Visible Surface Discrepancy (VSD). For more information on these metrics, please refer to \cite{sundermeyer2023bop}.

\noindent \textbf{Implementation Details:} We train the model on a single Titan XP GPU using Pytorch \cite{paszke2019pytorch} and Pytorch3d \cite{ravi2020accelerating} libraries. Our approach involves utilizing the same detection module and zoom-in strategy recommended in \cite{li2019cdpn,park2022dprost,wang2021gdr} to crop and resize the input query image of size $256 \times 256$, and using a size of $64 \times 64$ for geometric features for both query and references.

\begin{figure}[t]
  \centering
   \includegraphics[width=0.9\columnwidth]{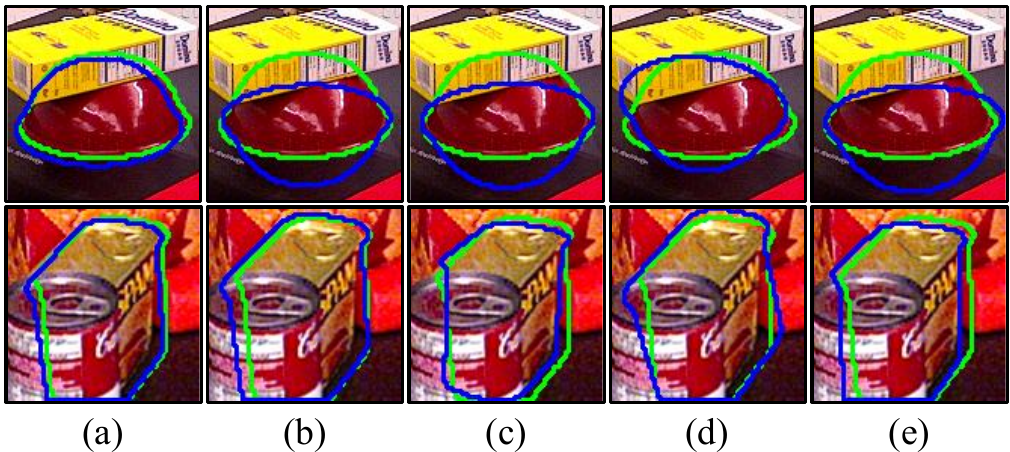}
   \caption{\textbf{Comparison with other SOTA methods:} We compare our method with others: (a) our method, (b) ZebraPose \cite{su2022zebrapose}, (c) CIR \cite{lipson2022coupled}, (d) GDRNPP \cite{wang2021gdr}, and (e) PFA \cite{hu2022perspective}. The projected contours derived from the label pose and the predicted pose are depicted in green and blue, respectively.}
   \label{fig:qualitative_sota}
\end{figure}

We evaluates the performance of the geometric feature extractor's encoder on two different scenarios: ResNet34 \cite{he2016deep} and ConvNeXT-Tiny \cite{liu2022convnet}, to compare the results of both a conventional model and a relatively recent model, which are computationally similar. The geometric feature extractor's decoder consists of a seven-layer CNN with two upsampling layers. Each pose refinement model comprises a three-layer CNN, two-layer FC layers, and two separate FC branches for translation and rotation estimation.

To train the model, we use the AdamW optimizer \cite{loshchilov2017decoupled} with a learning rate of 0.0002, cosine annealing with a cycle of 20k, and $N=5$ as the default frequency for positional encoding. We also use $M=4$ as the number of references.

In our experiment, we conduct four iterative single reference refinements and train the model for up to 600k steps to maximize the performance. However, in ablation studies of parameter setting, we train up to 100k steps with real data only for fast experiments.

\begin{figure*}[t]
  \centering
   \includegraphics[width=0.95\linewidth]{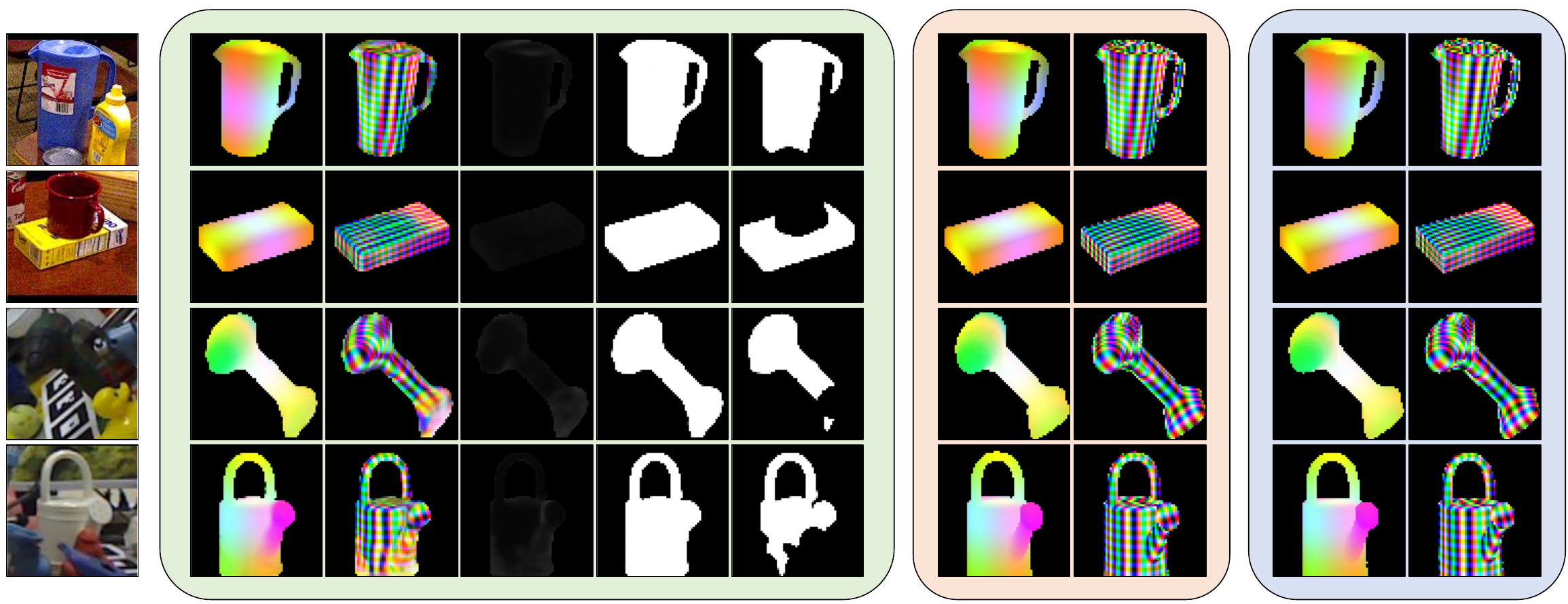}
   \caption{\textbf{Qualitative results from each module:} The first row displays the target image. In the green box, we show predictions from the geometric feature extractor, including low-frequency and high-frequency geometric features, average error, amodal mask, and modal mask, presented column-wise. The orange box presents the estimated pose from the multi-reference pose estimator, while the blue box shows the final pose estimations.}
   \label{fig:qualitative_all}
\end{figure*}

\begin{table*}[t]
\centering
{\resizebox{0.85\linewidth}{!}
{\begin{tabular}{c|c|cccc|cccc|c}
\hline
\multirow{2}{*}{Method}                                     & \multirow{2}{*}{Mean Time} & \multicolumn{4}{c|}{LM-O (pbr+real)}                                             & \multicolumn{4}{c|}{YCB-V (pbr+real)}                                         & \multirow{2}{*}{Mean AR} \\ \cline{3-10}
                                                            &                            & MSPD                 & MSSD              & VSD               & AR                & MSPD              & MSSD                 & VSD               & AR                &                          \\ \hline
GDRNPP \cite{wang2021gdr}                                   & 0.083                      & 0.875                & 0.664             & 0.518             & 0.686             & 0.813             & 0.799                & 0.694             & 0.769             & 0.727                    \\
Ext. PFA \cite{hu2022perspective}                           & 2.375                      & \underline{0.877}    & 0.761             & 0.597             & \underline{0.745} & 0.856             & 0.809                & 0.751             & 0.806             & 0.775                    \\
CIR \cite{lipson2022coupled}                                & -                          & 0.831                & 0.633             & 0.501             & 0.655             & 0.852             & \underline{0.835}    & 0.783             & 0.824             & 0.739                    \\
SCFLow  \cite{hai2023shape}                                 & -                          & -                    & -                 & -                 & -                 & \underline{0.860} & 0.778                & \textbf{0.840}    & \underline{0.826} & -                        \\ \hline
\begin{tabular}[c]{@{}c@{}}Ours\\ (Res34)\end{tabular}      & 0.093                      & \textbf{0.886}       & \textbf{0.769}    & \textbf{0.603}    & \textbf{0.753}    & 0.827             & 0.811                & 0.775             & 0.804             & \underline{0.778}        \\
\begin{tabular}[c]{@{}c@{}}Ours\\ (Convnext-T)\end{tabular} & 0.099                      & 0.875                & \underline{0.757} & \underline{0.598} & 0.744             & \textbf{0.873}    & \textbf{0.857}       & \underline{0.798} & \textbf{0.843}    & \textbf{0.799}           \\ \hline
\end{tabular}}
\caption{\textbf{Comparison on BOP Metrics:} We compare our approach with single model based methods using the MSPD, MSSD, and VSD metrics on both LM-O and YCB-V datasets.}
\label{table:BOP_metric}}
\end{table*}

\subsection{Comparison with State-of-the-arts}
In our study, we compared our approach with various state-of-the-art (SOTA) methods on the LM, LM-O, and YCB-V datasets. Recent methods can be divided into two groups: one group can estimate the pose directly from a zero-base, while the other group requires an initial pose estimator to narrow down the search space of their refinement module. Although the latter group often performs better, their dependence on the initial pose estimator and computational burden are inherent drawbacks. Therefore, we chose a direct pose estimation scheme without an initial pose estimator. 

To ensure scalability and practical learning, we implemented a strategy involving a unified model training scenario for all objects within each dataset. As a result, for a fair comparison, we exclusively assessed our method against approaches that similarly rely on a single model for all objects in each dataset.

\begin{table}[t]
\centering
    \resizebox{0.85\linewidth}{!}{
        \begin{tabular}{c|cc|cc}
            \hline
            Method               & \begin{tabular}[c]{@{}c@{}}DProST \\ \cite{park2022dprost}\end{tabular} & \begin{tabular}[c]{@{}c@{}}DProST\\ \cite{park2022dprost}\end{tabular} & Ours          & Ours                      \\ \hline
            M                    & 4                                                                       & 8                                                                      & 4             & 8                         \\ \hline
            ADD(-S) 0.02d        & 44.1                                                                    & 48.1                                                                   & \textbf{63.1} & \textbf{63.1}             \\
            ADD(-S) 0.05d        & 84.2                                                                    & 85.8                                                                   & \textbf{92.5} & \textbf{92.5}             \\
            ADD(-S) 0.1d         & 97.5                                                                    & 97.7                                                                   & 98.9          & \textbf{99.1}             \\ \hline
        \end{tabular}
    }
    \caption{\textbf{Comparison on mesh-less setting:} We compare our approach in mesh-less setting as suggested in \cite{park2022dprost}. The M is the number of references used for carving the object feature.}
    \label{table:mesh_less_sota}
\end{table}

Tab. \ref{table:LM_sota} presents the quantitative evaluations on the LM dataset. Our method outperforms most previous SOTA approaches. Notably, our method performs better than SCFlow \cite{hai2023shape} on the ADD(-S) 0.05d metric and the ADD(-S) 0.02d metric, which are more stringent than the ADD(-S) 0.1d metric. Moreover, we provide a detailed performance comparison of each object on the LM-O dataset. As shown in Tab. \ref{table:LM-O_sota}, our method outperforms SOTA methods under both settings, with ResNet34 and ConvNext-Tiny backbones.

Tab. \ref{table:YCB-V_sota} presents a comparison of the performance of the YCB-V dataset. Our method performs better than existing SOTA methods in both backbone settings. Even in challenging cases, such as those with occlusions and textureless, round bowls, our method accurately estimates the pose. Fig. \ref{fig:qualitative_sota} shows a visual comparison of our method with other existing methods, confirming that our method successfully overcomes the local minimum that the other method fails to address.

Moreover, in Fig. \ref{fig:qualitative_all}, we presents a comprehensive visualization of the qualitative results obtained from each module. The proposed model can successfully estimate the low and high-frequency positional encoding of the query object in the feature extraction module. Additionally, the refinement module demonstrates robust performance, even in the presence of severe occlusions, by leveraging the modal and amodal mask predictions.

We also benchmark our method against the methods listed on the \cite{sundermeyer2023bop}, where the listed methods have been optimized to perform better than their original papers. As shown in Tab. \ref{table:BOP_metric}, our ConvNexT-Tiny-based model outperforms the reported methods regarding Mean AR score while maintaining a fast inference time.

Furthermore, we conduct the experiment in a mesh-less setting by replacing the mesh with space-carved features, as suggested by \cite{park2022dprost}, and report the results in Tab. \ref{table:mesh_less_sota}. The results demonstrate that our method works accurately even in a mesh-less environment and outperforms \cite{park2022dprost} significantly.

\subsection{Ablation Study}

\begin{table}[t]
    \centering
    \begin{subtable}{.56\linewidth}
        \raggedleft
        \resizebox{\linewidth}{!}{
            \begin{tabular}[t]{c|ccc}
                \hline
                \multirow{2}{*}{Type of $G$} & \multicolumn{3}{c}{ADD(-S)}                   \\
                                                & 0.02d         & 0.05d         & 0.10d         \\ \hline
                (R, G, B)                       & 48.4          & 84.2          & 96.5          \\
                (X, Y, Z)                       & 49.9          & 86.4          & 97.4          \\
                (X, Y, Z, d)                    & 50.4          & 86.5          & 97.5          \\
                $\gamma$5(X, Y, Z)                     & \textbf{56.0} & \textbf{89.1} & \textbf{98.1} \\
                $\gamma$5(X, Y, Z, d)                  & 53.7          & 87.8          & 97.7          \\ \hline
            \end{tabular}
        }
        \subcaption{}
    \end{subtable}
    \begin{subtable}{.42\linewidth}
        \raggedright
        \resizebox{\linewidth}{!}{
            \begin{tabular}[t]{c|ccc}
                \hline
                \multirow{2}{*}{N} & \multicolumn{3}{c}{ADD(-S)}                   \\
                                                & 0.02d         & 0.05d         & 0.10d         \\ \hline
                3                     & 54.5          & 88.7          & 97.7          \\
                5                     & \textbf{56.0} & \textbf{89.1} & \textbf{98.1} \\
                7                     & 54.6          & 88.5          & 97.9          \\ \hline
            \end{tabular}
        }
        \subcaption{}
    \end{subtable}
    \caption{\textbf{Ablation studies I} (a) \textit{Geometric features:} We compare different geometric feature types for the query and references. (b) \textit{Number of frequencies:} We evaluate performance for the number of frequencies in positional encoding.}
    \label{table:ablation_geometric}
\end{table}

\begin{figure}[t]
  \centering
   \includegraphics[width=0.85\columnwidth]{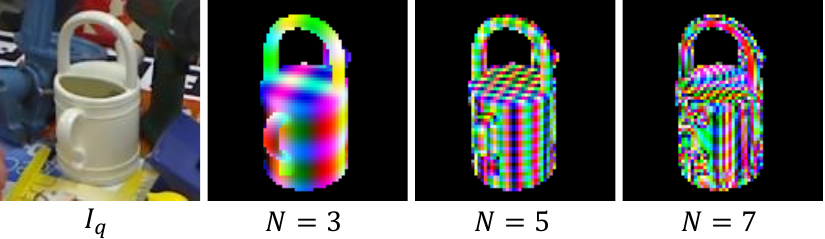}
   \caption{\textbf{Qualitative results of the positional encoding frequency:} We visualize the geometric features across different frequencies.}
   \label{fig:qualitative_freq}
\end{figure}

\noindent \textbf{Effectiveness of Geometric Representations:} We first analyze the effectiveness of our proposed geometric feature representation, based on positional encoding, compared to existing methods. As shown in Tab. \ref{table:ablation_geometric}(a), our method outperforms the RGB projection used in render-and-compare strategies \cite{li2018deepim, labbe2020cosypose, park2022dprost}, and XYZ values of the 2D-3D coordinates used in \cite{wang2021gdr, park2019pix2pose}. Additionally, we explore the potential benefits of integrating the normalized camera-to-object ray direction as presented in \cite{mildenhall2020nerf}. However, we have found that although using rays enhances the performance without positional encoding, our model with positional encoding performs better without rays. Therefore, we have decided not to include rays in our final geometric representations.

We investigate the optimal frequency embeddings for positional encoding in Tab. \ref{table:ablation_geometric}(b). We observe that performance peaks at $N=5$ and declines after that. This decrease is likely due to aliasing at higher frequencies, as depicted in Fig. \ref{fig:qualitative_freq}, leading to a degradation in performance.

\begin{table}[t]
    \centering
    \begin{subtable}{.4\linewidth}
        \raggedleft
        \resizebox{\linewidth}{!}{
            \begin{tabular}{c|ccc}
                \hline
                \multirow{2}{*}{M} & \multicolumn{3}{c}{ADD(-S)}                   \\
                                                & 0.02d         & 0.05d         & 0.10d         \\ \hline
                1                               & 54.7          & 88.3          & 97.9          \\
                4                               & 56.0          & 89.1          & \textbf{98.1} \\
                8                               & \textbf{56.3} & \textbf{89.3} & \textbf{98.1} \\ \hline
            \end{tabular}
        }
        \subcaption{}
    \end{subtable}
    \begin{subtable}{0.56\linewidth}
        \centering
        \resizebox{\linewidth}{!}{
            \begin{tabular}[t]{c|ccc}
                    \hline
                    \multirow{2}{*}{Setting} & \multicolumn{3}{c}{ADD(-S)}                  \\
                                             & 0.02d        & 0.05d         & 0.10d         \\ \hline
                    w/o Occ. Aug.            & 6.2          & 35.1          & 66.3          \\
                    w/o AdaIN                & \textbf{7.9} & 38.9          & 67.8          \\
                    Ours                     & 7.5          & \textbf{39.6} & \textbf{68.2} \\ \hline
            \end{tabular}
        }
        \subcaption{}
    \end{subtable}
    \caption{\textbf{Ablation studies II} (a) \textit{Number of references:} We check the performance with the number of references M. (b) \textit{Addressing occlusions:} We evaluate the strategies for occlusions, including occlusion augmentation and the AdaIN on the LM-O dataset.}
    \label{table:ablation_2}
\end{table}

\noindent \textbf{Ablation on references:} We assess our model's performance under different reference settings. Specifically, we experiment with the optimal number of references per object, as presented in Tab. \ref{table:ablation_2}(a). Since the performance saturates when using more than four references, we use four as our default setting.

\noindent \textbf{Ablation on occlusion addressing strategy:} Tab. \ref{table:ablation_2}(b) shows the performance enhancements from our occlusion augmentation and the AdaIN method in the geometric feature extractor.

%% file: sec/5_conclusion.tex
\section{Conclusion}
This paper discusses the problems related to blurry geometric estimation and local minimums commonly faced in traditional pose estimation techniques. To overcome these issues, we propose a new method that directly estimates positional encoding values of the object surface. This approach helps in multi-reference refinement, further improving the pose estimation's accuracy. In order to validate our approach, we have conducted extensive experiments on different types of geometric features and introduced an update algorithm independent of the intrinsic matrix. Additionally, we have developed an AdaIN module and an augmentation method that aim to reduce performance degradation caused by occlusions. Experimental results demonstrate that our method performs superior on the LM, LM-O, and YCB-V datasets and outperforms existing methods in a mesh-less setting. Our plans involve incorporating additional geometric representations to further enhance performance based on our findings that elaborate geometric features provide advantages.